# Biologically Inspired Design Principles for Scalable, Robust, Adaptive, Decentralized Search and Automated Response (RADAR)

Melanie Moses and Soumya Banerjee
University of New Mexico
Albuquerque, New Mexico, USA
melaniem@unm.edu

*Abstract*— **Distributed search problems are ubiquitous in Artificial Life (ALife). Many distributed search problems require identifying a rare and previously unseen event and producing a rapid response. This challenge amounts to finding and removing an unknown needle in a very large haystack. Traditional computational search models are unlikely to find, nonetheless, appropriately respond to, novel events, particularly given data distributed across multiple platforms in a variety of formats and sources with variable and unknown reliability. Biological systems have evolved solutions to distributed search and response under uncertainty. Immune systems and ant colonies efficiently scale up massively parallel search with automated response in highly dynamic environments, and both do so using distributed coordination without centralized control. These properties are relevant to ALife, where distributed, autonomous, robust and adaptive control is needed to design robot swarms, mobile computing networks, computer security systems and other distributed intelligent systems. They are also relevant for searching, tracking the spread of ideas, and understanding the impact of innovations in online social networks. We review design principles for Scalable Robust, Adaptive, Decentralized search with Automated Response (Scalable RADAR) in biology. We discuss how biological RADAR scales up efficiently, and then discuss in detail how modular search in the immune system can be mimicked or built upon in ALife. Such search mechanisms are particularly useful when components have limited capacity to communicate and when physical distance makes communication more costly**.

## I. INTRODUCTION

Search and response processes are ubiquitous in Artificial Life (ALife). Distributed systems can use effective search and response to coordinate a multi-robot control system, search for and acquire resources in a distributed peer-to-peer system, or detect and prevent a cyber attack in progress. Many of these tasks require identifying a rare and previously unseen event and producing a rapid response. This challenge amounts to finding and removing an unknown needle in a very large haystack. Traditional computational search models are unlikely to find, nonetheless, appropriately respond to, novel events, particularly given data distributed across multiple platforms in a variety of formats and sources with variable and unknown reliability.

Biological systems have evolved solutions to distributed search and response under uncertainty. When faced with a lethal pathogen, the immune system must rapidly find and neutralize a small number of foreign pathogens hiding among trillions of healthy host cells, or the host dies. Similarly, ant colonies use decentralized search and control to exploit new resources and defend their territories, making them dominant players in ecosystems across the globe [1].

Immune systems and ant colonies efficiently scale up massively parallel searches and responses in highly dynamic environments, and both do so using distributed coordination without centralized control. Millions of ants in a colony and trillions of cells in the immune system parallelize the search for food or pathogens with apparently little communication overhead as systems increase in size. These properties are relevant to computer science, where distributed, autonomous, robust and adaptive control networks are important in the design of robot swarms, mobile computing networks and other distributed intelligent systems.

Immune systems have evolved very complex mechanisms to rapidly solve very difficult search problems. The primary adaptive immune cells that recognize and neutralize pathogens are lymphocytes called B cells and T cells (also called white blood cells). Each lymphocyte can bind to a unique set of proteins called 'cognate' antigens on the surface of viruses, bacteria or infected host cells.

Collectively, the population of lymphocytes can recognize on the order of $10^6$ different antigens [2]. Lymphocytes interact with dozens of other cell types and chemical information signals called cytokines. Here, we focus on how the architecture of the lymphatic network facilitates the search for pathogens and production of antibodies by B cells.

Ant colonies have also developed a wide diversity of strategies for effective search among large numbers of individuals connected by a distributed communication network. The behavior of an ant colony is not directed by any individual ant; rather, it emerges from interactions between individuals. Individuals communicate by laying chemical pheromone trails to recruit nest mates to food, fights or other tasks. Individuals also interpret the rate of encounter with other individuals as information signals [3], for example, workers wait safely inside the nest until foragers return at a rate above a threshold that indicates that there is abundant food. The complexity of information exchange increases with colony size—larger colonies solve more difficult search problems by inventing new modes of communication in real ants [4] and in mathematical models [5].

Immune systems and ant colonies have inspired many successful computational algorithms. Immune system inspired approaches have been particularly successful in computer security (reviewed in [6]), and ant colony inspired approaches provides effective heuristics for searching dynamic environments (reviewed in [7,8]) and routing (e.g. [9]). These methods are distributed, scalable and sometimes robust to small failures, but there has been little success in scalable automated response (but see [10]).

Here, we elucidate how rapid decentralized response is built into biological systems. We characterize the design principles in ant colonies and immune systems that lead to scalable RADAR (Robust Adaptive Decentralized search Automated Response). We show how innovation in communication pushes the scaling boundaries to allow systems to grow. Finally, we suggest a roadmap for incorporating automated response into the next generation of biologically inspired methods to achieve Scalable RADAR in ALife.

The paper is organized as follows. In section 2, we review design principles for Scalable RADAR in ant colonies and immune systems, and discuss how biological RADAR scales up efficiently. In section 3 we discuss systematic changes in systems as they grow and theoretical limits to size. In section 4, we discuss how innovations in communication facilitate decentralized search. In section 5 we demonstrate how the immune system uses semi-modular search to balance tradeoffs between local and global communication. In section 6 we apply scalable RADAR principles to ALife problems.

II. SCALABLE RADAR SEARCH PRINCIPLES IN ANT COLONIES AND IMMUNE SYSTEMS

Ant colonies and immune systems are canonical complex adaptive systems with decentralized control. Despite some differences in scope, scale and complexity, there are general principles that guide RADAR in both systems. Search in these systems is characterized by the following attributes.

1) **Robust**: Redundancy, flexible diversity, and probabilistic response to partial information all contribute to robustness. Redundancy of components means that if one cell or ant dies, there are multiple similar ants or cells to take its place. There is diversity among components, but the exact role or task of a component is flexible. Individuals are divided into task groups and cell types. If, for example, the foraging task group is not large enough, maintenance workers notice the slowing rate of incoming food and switch tasks to begin foraging. Large numbers of individuals respond probabilistically to information signals. Many responses to information signals include amplifying the original signal. While sometimes this can lead to positive feedbacks that grow out of control (i.e. cytokine storms in response to influenza infection [11]), generally the 'wisdom of the crowd' [12] is effective even when individuals make mistakes. Only when a large number of individuals respond inappropriately to a signal does positive feedback lead to an erroneous response by the whole system.

2) **Adaptive**: Individuals or populations of individuals adapt in response to environmental signals. For example, when an ant discovers a new food source and lays a chemical pheromone trail to that food source, other ants change their behavior by following the trail with some probability. There is a diversity of thresholds in the population of ants so that some follow small concentrations of pheromone and others respond only to larger pheromone concentrations. When a B cell binds to a pathogen and is activated by other immune system cells, that B cell produces a large and variable population of daughter B cells. Those that bind to the pathogen most effectively reproduce faster, so the population of cells improves its ability to neutralize the pathogen.

3) **Decentralized search**: No ant or cell tells the other ants or cells what their task is, or when or where they should do it. Yet both systems search dynamic and complex landscapes effectively. Individuals sense chemical signals (pheromones or cytokines), rates of interaction with environmental cues or other individuals to determine how, when and where to search. While control is completely decentralized, communication between individuals is aggregated spatially, e.g. lymph nodes and branches on foraging trails concentrate interactions and chemical signaling between individuals.

4) **Automated Response**: The response of colonies and immune systems is as distributed as search. Individuals act (e.g., kill an infected cell, move along a chemical gradient, release chemical signals) by integrating local signals from their environment to determine their behavior. Some local responses may be 'errors' (e.g. choosing the wrong food or killing a healthy cell), but actions only proliferate through the whole system when positive feedbacks are generated by many individuals that have the same response to stimuli. Thus individuals may make mistakes, but the whole system response is governed by collective agreement.

5) **Scalable**: Ant colonies consist of tens to millions of ants, and immune systems may consist of trillions of cells. Because all ants and cells respond only to local signals, each can act in parallel without need for information signals to travel to every individual for a search to be effective or a response to be initiated. However, we posit below that scalable response requires more connections between individuals as the system grows, and that responses of individuals likely vary with system size. Larger colonies have a greater diversity of castes and task groups with more specialization, are more likely to communicate and recruit to collect food, have more complex nest structures, and have more sophisticated processes to exploit resources and compete with other colonies. In comparison, immune systems are vastly more complex, with many more cell types, signaling molecules and specialized receptors to interpret them. The immune system of an elephant produces millions more antibodies than the immune system of a mouse, but in approximately the same time.

There are some clear differences between search in ant colonies and immune systems. Immune systems have orders of magnitude more components, many more cell types and signaling molecules, and more complex networks of interaction among them. Some of the complexity of the immune system is driven by the need of the immune system to recognize and avoid attacking self—autoimmune diseases are clear examples of failed automated response. Additionally the immune system searches for rapidly moving targets: pathogens can mutate to avoid immune system recognition, for example, in HIV mutates so rapidly that an entirely new genetic strain evolves in the course of a single infection [13]. Pathogens are much more sophisticated about avoiding the immune system than seeds are at avoiding ants, and so, the immune system has evolved more sophisticated scalable RADAR.

While some aspects of immune response and ant foraging are well studied, many questions remain about the basic biology of these systems. For example, it is not clear how automated responses are turned off, e.g. shutting down immune response, or ending a fight with a competitor or abandoning a food source. It is not known what kinds of individual mistakes are likely to persist to affect behavior of the whole system. Sometimes the immune system attacks itself or an ant colony chooses a clearly inferior nest location. Why these mistakes happen, and whether, what kind and how many mistakes are inevitable for distributed search remains an open question.

Particularly relevant to this paper, few details are known about how the architecture of the nest, foraging trails, or lymphatic network facilitates efficient contacts and exchange of signals among cells. We outline what is known below. We hypothesize that scalability is promoted by 1) physical architectures of lymph nodes and lymphatic ducts or ant nests and foraging trails, and 2) innovations that allow more interactions and communication between individual components as the system grows.

### III. THEORETICAL LIMITS TO SCALING RADAR

In order to understand scaling limits to decentralized networks, we turn first to the better-understood scaling limits of centralized networks. Theory shows that centralized networks that distribute energy or materials are characterized by diminishing returns which cause them to be less efficient as they increase in size [14,15,16]. We hypothesize that decentralized networks avoid much of the overhead of centralized control that causes diminishing returns. However, we understand only a few of the mechanisms that ant colonies and immune systems use for scalable decentralized coordination.

Metabolic Scaling Theory [14,17] explains why there are diminishing returns in centralized networks like the cardiovascular system. The cardiovascular network delivers energy to each cell at a slower rate in large animals. Specifically, metabolic rate, $B$, scales with mass, $M$, following $B = cM^{3/4}$, where $c$ is a constant specific to a taxonomic group, and ¾ is a scaling exponent that indicates that a 100-fold increase in mass leads to only a 30-fold increase in metabolism. Since the scaling exponent is usually of primary interest, the scaling relationship is usually written as

$$B \sim M^{3/4} \quad (1)$$

The theory posits that the ¾ exponent arises from the fractal-like branched geometry of the cardiovascular network. Because blood flows a longer distance in a larger organism, it takes more time to travel through the branched arteries, so that circulation times scale proportional to $M^{1/4}$ and heart rates scale as $M^{-1/4}$. As a result, a very long list of biological rates are slower in larger animals: lifespan, gestation time, growth rate, reproductive rate, heart rate, and even rates of evolutionary change across generations are characterized by ¼ power scaling, all because of the diminishing returns in centralized networks [17].

In addition to explaining why mice live fast and die young, the theory sets a limit to how large organisms can grow. Animal growth can be modeled as a function of metabolic intake and metabolism allocated to maintenance, leading to the growth equation

$$dm/dt = am^{3/4} - bm \quad (2)$$

where m is mass at time $t$, $a$ is the metabolic energy required to generate a unit of biomass and $b$ is the power needed to maintain a unit of biomass [18,19]. Maximum mass, $M$, can be found by setting $dm/dt = 0$, so that $aM^{3/4} = bM$, or $M = (a/b)^4$. Thus, the centralized flow of resources and basic energetic properties of cells set the maximum size of organisms.

In contrast to diminishing returns in centralized distribution networks, decentralized interaction networks may be characterized by increasing returns, for example, interaction rates increase in larger cities [20] and larger ant colonies [21,22,23]. Metcalf's law hypothesizes that the value of a network increases proportional to $n^2$, where $n$ is the number of nodes in the network, although that theory may be too optimistic [24]. The superlinear scaling reflects the increasing number of interactions that can lead to information exchange. Each of $n$ nodes in a network can contribute information to every other $n-1$ nodes in the network. Thus, in information networks, increasing size leads to more potential interactions per node (with a per node exponent of 1), as system size grows, as opposed to decreasing flow of material or energy per node (or per cell) in centralized systems (with a per cell exponent of -1/4).

Ant colonies and immune systems are characterized by centralized material flow and decentralized information exchange. For example, ants build trail networks to gather food from the environment and transport it back to a central nest, which means more transport through larger territories, and lymphocytes travel longer distances through the circulatory network in larger animals. Additionally, these systems are composed of interaction networks among ants or cells, and greater numbers of individuals can interact in more ways, potentially leading to increasing amounts of information exchange. Thus, these systems achieve some balance between decentralized interactions and centralized flow.

If we consider the size, $n$, of a colony or immune system, we expect that flow of materials or energy to scale as $n^\alpha$, where $\alpha$ is less than 1 (3/4 in a 3 dimensional immune system, but 2/3 in a primarily 2 dimensional foraging network). We expect the number of potential interactions to scale as $n^\beta$, where $\beta = 2$ according to Metcalf's law, although the exponent may increase, if for example, the number of signaling molecules increases with size, or decrease if the ability to interact saturates. Following, Bettencourte et al [20], if interaction rates lead to innovations that increase energy flow or reduce time, for example if information exchange among foragers allows foragers to find better sources of food, then metabolic intake would scale as $n^\gamma$, where $\gamma$ is an exponent larger than 2/3, and depends on how the rate of interactions ($n^\beta$) alters the rate of metabolism. Bettencourte considers the effect of such modifications to scaling theory to explain how cities continue to grow to larger sizes. While there are diminishing returns in moving resources through larger cities, cities are engines of innovation, and grow according to $dn/dt = an^\gamma - bn$ where now $\gamma$ can be greater than 1, meaning that there is no limit to city size as long as innovations continue to increase with size. Bettencourte shows that the interplay between diminishing and increasing returns leads to boom and bust cycles, and the rate of innovation must continue to increase to keep the system from collapsing.

Empirically, ant colonies and immune systems conform to some ¼ power scaling patterns, but not others. For example, immune response times appear to be mass-invariant, or very nearly so: the immune system of a horse responds to a pathogen as quickly as the immune system of a sparrow [25,26,27]. There are obvious evolutionary pressures that demand fast immune response in large animals, since ¼ power scaling would turn the 3 day immune response of a sparrow into a month in a horse, long after the pathogen had killed it. On the other hand, immune memory of past pathogen exposures lasts a lifetime, proportional to $M^{1/4}$. Ant colonies appear to escape diminishing returns in foraging rates, with large colonies foraging at the same per ant rate as small colonies, despite having to transport food further [23, 31, 32]. But, the metabolism of entire colonies increases with an exponent close to ¾ [28,29] and some aspects of colony reproduction follow 1/4 power scaling predictions [30].

Although decentralized search and response may avoid the diminishing returns of centralized systems, coordination that produces coherent and robust response is particularly difficult when systems are big. As systems grow, innovations are required to enable decentralized coordination of search and response. For example, the immune system of an elephant has to find the same small number of viruses injected by a mosquito vector in a space that is a million times larger than that of a mouse. It also has to produce a million times more antibody in the same few days in response to that infection. A purely modular approach in which a population of lymphocytes patrolled a local area of tissue could produce fast scalable search; however, proliferation of antibodies in the local lymph node is not sufficient to produce a high enough concentration of antibodies in the blood [26,27]. Thus, local detection has to trigger a global response of appropriate B cells to produce antigen. We hypothesize that the architecture of lymph nodes and circulatory networks in the immune system balances fast scalable modular search with rapid global response to 'think locally but act globally.'

We hypothesize that physical architectures, diversity and specialization of components, signaling and information processing enable biological systems to scale up over many orders of magnitude. In the following sections we show that ant colonies evolve increasingly sophisticated innovations in communication as colony size increases, and that the physical architecture of the immune system promotes communication and minimizes response times as the system increases in size.

## IV. INNOVATIONS IN COMMUNICATION PUSH THE SCALING BOUNDARIES IN ANT COLONIES

Ant colony size varies by over a million-fold, and the number of cells in immune systems are orders of magnitude larger than that. There are inherent costs to large size: it is harder to search larger spaces, to transport individuals and information signals farther, and to maintain a coherent response in more individuals. There are also inherent benefits to large size—more components allows for a greater diversity with more specialized roles, a greater variety of response thresholds, and more ability to sample the environment. The crowd is wiser when it is larger, but only if local solutions can aggregate into effective global responses.

Large size is often accompanied by innovations that allow components to be more connected. Larger colonies use more pheromone signaling, and individual ants likely respond more quickly to interactions with nest mates. Computational models of pheromone mediated foraging assume that the probability of foraging in a direction is proportional to the amount of pheromone in that direction:

$$p_{ij} = \frac{\tau_{ij}^\alpha}{\sum_k \tau_{ik}^\alpha} \qquad (3)$$

where $p_{ij}$ represents the probability of traveling from node $i$ to node $j$, given an amount of pheromone $ij$ along that route, raised to an arbitrary exponent $\alpha$, divided by the summed pheromone on all other available paths $k$.

Typically, pheromones are laid by ants traveling from a food source back to the nest. The pheromone biases the next ant leaving the nest to follow that trail, and by the second ant reinforcing the pheromone trail, a positive feedback is setup whereby more ants choose to travel toward food and they recruit still more ants as long as food remains. The pheromones decay over time, so if food is depleted, the trail dissipates and ants choose other directions to travel. This mechanism is particularly effective when there are many ants, both because the positive feedback is reinforced rapidly, and because there can be a greater diversity of values of $\alpha$. Ants with low values of $\alpha$ will ignore pheromone trails and continue to explore for new food sources, while ants with high $\alpha$ exploit known resources. When there are more ants in the colony, each ant communicates with more other ants by laying pheromone, but with the same amount of effort [32]. We hypothesize that large colonies are effectively 'smarter' than small colonies [4,8] and greater use of information compensates for the long distances that large colonies must transport food [23,31].

Ants in the largest colonies may travel 50 meters to find food, while the smallest colonies may live inside a single fruit. However, our experimental studies show that large seed harvesting colonies collect food on a per capita basis just as fast as small colonies, even though they spend significantly more time transporting food [23,31,32]. The large colony also lives in a more variable world, where identifying rich food patches can counteract the significant travel and transport problems that come with larger size. We hypothesize that increased communication in larger colonies counteracts the costs of larger transport and more difficult search in larger colonies. Another mechanism by which increased communication is facilitated by large colonies arises because larger ant colonies are more crowded: ants have more opportunity to interact and exchange information, and every ant benefits from the signals produced by every other ant. With a million ants searching, the best locations are more likely to be found, and improved communication allows every ant to benefit from every new discovery.

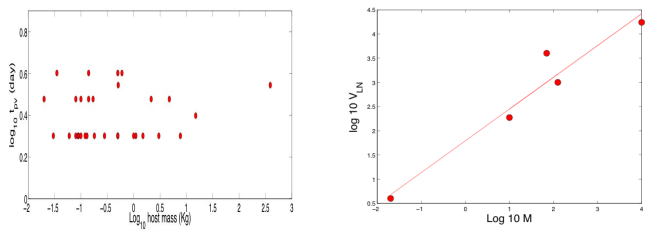

Figure 1. (Left Panel) Empirically measured time between infection and peak viral concentration in blood ($t_{pv}$) vs. host body mass ($M$) on logged axes (there is no significant slope, p = 0.35, reproduced from [27]) for species ranging from sparrows to horses. (Right Panel) Volume of a lymh node ($V_{LN}$) vs. host body mass on logged axes (there is a significant slope with p = 0.006 showing a clear increase with mass as predicted by Eq. 4).

## V. SEMI-MODULAR SEARCH IN THE IMMUNE SYSTEM

We define semi-modular search in the immune response to West Nile Virus (WNV), which is a generalist virus that infects many vertebrate hosts and has caused substantial epidemics in north American birds since its introduction into North America in 1999 [33]. A constant number (approximately $10^5$) of WNV virions are injected into a host by a mosquito, regardless of host size [34]. The immune system in bigger organisms can find this fixed number of virus particles in approximately the same time as in smaller organisms (Fig. 1 Left Panel, data for species from sparrows to horses), i.e. the immune system of a horse searches a volume of tissue 10,000 times larger than a sparrow, but in the same time. This rapid search is not surprising since the horse has 10,000 more copies of lymphocytes than the sparrow. However, in order to neutralize the pathogen, the horse has to produce 10,000 times more antibody than the sparrow, again in the same time, which empirically for WNV is approximately 3 days. The B cells that recognize the WNV antigen proliferate exponentially, doubling their numbers approximately twice per day, to produce a population of antibody producing plasma cells. In order to produce 10,000 times more antibody in a fixed time period, the horse has to activate 10,000 times more of the B cells that recognize WNV.

Thus, relatively uncommon B cells specific to WNV must locate initially rare antigen localized in a small region of tissue and then respond by activating a large global population of similar B cells to produce sufficient antibody to circulate through the blood. The immune system has to think locally, but act globally.

The immune system achieves this balance by distributing immune cells across a semi-modular hierarchical system of lymph nodes connected by the circulatory and lymphatic networks. Each lymph node is a central location in a small 'draining region' of tissue. Dendritic cells carry antigen from the tissue to the draining lymph node, and dendritic cells loaded with potential pathogens search for cognate lymphocytes in the relatively small, spatially structured lymph notes. Search within each lymph node and its associated draining region occurs in parallel. It is thought that at least one copy of each type of lymphocyte circulates within each lymph node [35].

Empirically, lymph node size and the size of the draining region both increase with organism size, but the increase is sublinear [26,27]. Thus a horse has more and larger lymph nodes than a mouse, but neither the increase in lymph node size nor the increase in number are as great as the increase in body size. The larger size of the lymph nodes and draining regions in a horse means that the cells have to travel farther in order to encounter each other and the pathogen. We hypothesize that this causes search to take slightly longer in the horse, although the increase in time is not detectable within the once per day time frame in which measurements are taken.

Completely modular systems have no overhead of communication and hence achieve perfectly parallel search [36] since search is in a space of the same size and is replicated in parallel. However, the need for automated response makes complete modularity impractical—although the search for pathogens could happen locally and independently, the response to pathogens sometimes requires global communication: immune cells in different lymph nodes have to communicate in order to mobilize sufficient antibodies to fight the pathogen, and thus the number of B cells and lymph nodes that communicate increases with the size of the system. Thus, the need for global response influences the immune system architecture, and prevents a simpler modular design.

We can show an optimal architecture that minimizes the total time for lymphocytes to detect antigen locally but activate other lymphocytes to produce a global response. The time taken to carry antigen from the tissues in the draining region scales in proportion to the radius of that draining region (it is the time for dendritic cells to crawl from the site of a mosquito bite to the nearest lymph node, typically several hours in a mouse [37]). Since the draining region volume is proportional to lymph node volume ($V_{LN}$), that distance, and detection time ($t_{detect}$) scale as $t_{detect} \propto V_{LN}^{1/3}$. The response to the infection requires activating a number of lymphocytes proportional to $M$ (in order to produce a mass-invariant concentration of antibody in a fixed period of time). Activation occurs by recruiting lymphocytes into the lymph node at the site of infection, though high endothelial venules (HEV), the number of which is proportional to lymph node size.

The number of lymph nodes that a single infected site lymph node has to communicate with ($N_{comm}$) in order to recruit more B cells is proportional to the amount of antibody required to neutralize the pathogen divided by the number of B cells resident in a lymph node ($Num_{Bcell}$): $N_{comm} \propto M / Num_{Bcell}$. Noting that $Num_{Bcell} \propto V_{LN}$ (a larger lymph node will have more immune cells within it), we have $N_{comm} \propto M/V_{LN}$. The rate at which new B cells from other lymph nodes enter the infected site lymph node ($rate_{comm}$) is proportional to the number of HEVs which in turn is proportional to the size of the lymph node: $rate_{comm} \propto V_{LN}$. The time spent in communication is now $t_{comm} = N_{comm}/rate_{comm} \propto M/V_{LN}^2$. Minimizing the sum of $t_{detect}$ and $t_{comm}$ gives

$$V_{LN} \propto M^{4/7} \text{ and } N \propto M^{3/7} \qquad (4)$$

where $N$ is the number of lymph nodes and $V_{LN}$ is the volume of a lymph node [27] (Fig. 1 Right Panel).

Thus, lymph node numbers and sizes both increase sub-linearly with organism size. The semi-modular architecture reduces RADAR search and response times in the immune system to scale as $M^{1/7}$.

This analysis suggests that even in the largely decentralized immune systems, there are increasing costs to global communication as organisms grow bigger. The increasing cost is a result of the requirement to activate larger numbers of IS cells for antibody production in larger organisms. A semi-modular architecture (Fig. 2) balances the opposing goals of detecting antigen using local communication and producing antibody using global communication and leads to optimal antigen detection and antibody production time. Larger lymph nodes have more HEVs and hence can recruit faster (lower global communication cost) but since there are fewer lymph nodes, each is in charge of a larger volume of tissue. This implies that dendritic cells have to travel further to get to the nearest lymph node (larger local communication cost). Lymphocyte trafficking, the architecture of the lymph nodes, lymphatic network and circulatory network likely play a role in speeding up long distance communication, but these biological details are not yet well understood.

## VI. RELEVANCE OF SCALABLE RADAR TO ALIFE

The immune system demonstrates that while search can be modular and local, coordinated response requires communication between components. The immune system is designed such that larger organisms have larger lymph nodes. The larger size of those lymph nodes enables more 'long distance' communication since the number of HEV is proportional to lymph node size. Those HEV connect to the circulatory network, which allows transport of lymphocytes from distant lymph nodes in seconds or minutes, very fast compared to the slow crawl of dendritic cells through local tissue. The architecture of the immune system results in immune response times scaling up efficiently due to a greater capacity for fast global communication in larger systems.

Ant colonies demonstrate an increasing number of innovations in communication as system size increases. Ants go from individual foraging, to foraging with recruitment within a nest to information sharing between nests as colony size increases [4].

In both ant colonies and immune systems a successful search (a lymphocyte binding to a pathogen or an ant discovering a new seed pile) sets off a cascade of events that simultaneously increases the number of individuals searching in that area and initiates a response. The 'area' need not be physical space. For example, when a B cell detects a cognate antigen, it undergoes processes called somatic hypermutation and clonal selection. This generates new populations of B cells that are 'nearby' in antigen space (B cells that recognize similar but not identical antigens), and the B cells that bind most closely to the antigen proliferate preferentially. Thus, when a B cell encounters an antigen that it has some capacity to bind to, it creates new populations of B cells to search for similar antigen.

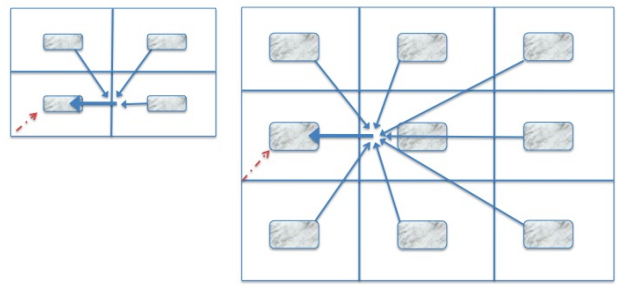

Figure 2. A semi-modular detection network of lymph nodes (LNs). The shaded regions are LNs and the unshaded regions are the DR. The hypothetical organism on the right is four times as big as the one on the left. The number of LNs and their size both increase with the size of the organism. The local communication (antigen loaded dendritic cell migration to LN) is shown by a dotted arrow and the global communication (recruitment of NIS cells from other LNs) is shown by solid arrows. The size of the incoming arrow into a LN represents the size of the HEV (which is proportional to the size of a LN).

These biological design principles--balancing exploitation of known resources and exploration of new ones (Fig. 3), network architectures that facilitate increased communication with increasing size, and innovations in communication capacity as size increases, can be incorporated into artificial systems that require scalable RADAR. Such systems include searching online social networks for information, aggregating data from multiple sources in peer-to-peer systems, distributed control of mobile robots, adaptive dynamic routing using Ant Colony Optimization and automated response to cyber attacks.

For example, a previous immune inspired work uses process homeostasis [10] to automatically respond to computer security attacks by slowing communication. The scalable RADAR principles suggest that slowing should occur in a small local region where the attack was detected, recruiting more sensors to that subnet, deploying a local response and increasing the scope of response only if the local response is successful.

The principles are also relevant for search in social networks. We argue that in social systems, scalable architectures and innovations in communication are required as the system grows. We can compare the approach by the immune system to how search scales over 'small world' networks, inspired by Stanley Milgram's classic experiment [38] in which individuals were instructed to forward letters to other unknown individuals using only information about their immediate acquaintances. Milgram found the counterintuitive result that communication could be established using very few connections (six degrees of separation) between individuals separated by great social and geographic distances. Understanding how short paths can be found through large graphs, using only local information, remains a theoretical and practical challenge [39].

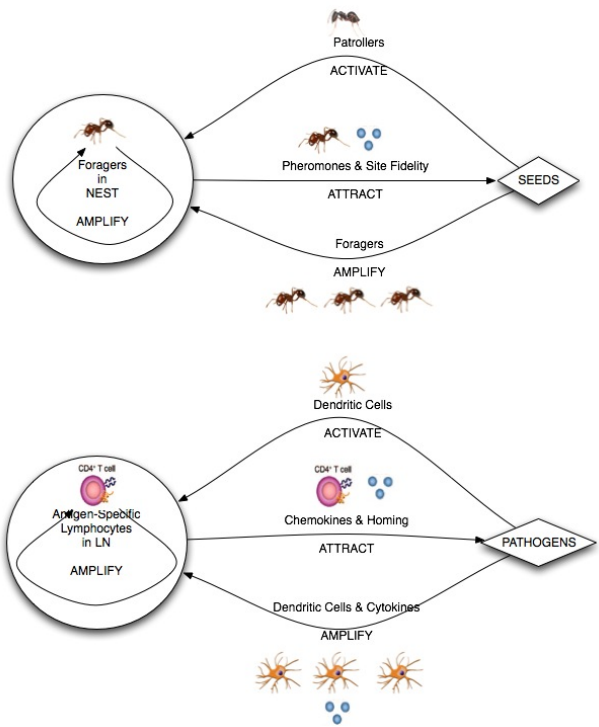

Fig. 3. Parallel interactions in ant foraging and immune response. In each system, one set of components (patrollers or dendritic cells) **activates** the search when they encounter seeds or pathogens. The components that actively search (foragers or lymphocytes) **amplify** their numbers. Foragers leave the nest in response to the return rate of other foragers and lymphocytes proliferate in response to activation by dendritic cells. Search may be guided by chemical signals (pheromones or chemokines) to **attract** ants or cells to the site where seeds or pathogens have been found.

Efficient scalable search over small world networks requires more communication between nodes as system size increases. This 'densification', appears as increasing out degree of nodes with the number of nodes in the small world network. Specifically, with a node out-degree $k = c(\log n)^2$ where $n$ is the number of nodes in the system, the expected delivery time of a message was proven to be $O(\log n)$ [39]. These results are in agreement with Milgram's original experiments [38] and lend insight into the puzzling riddle of "six degree's of separation" [40] in social networks. Analysis of empirical data on online social networks [41,42] also reveal that densification also facilitates efficient search in such networks. This process of densification is similar to what we found in ant colonies and immune systems: larger networks require more communication between individual components.

The advantages of scalable RADAR principles have been described for two engineered systems. Multi-robot control systems in which high memory and bandwidth computer servers direct power and bandwidth constrained mobile robots, are analogous to lymph nodes in charge of dendritic cells servicing the draining region [27]. Such systems can have tradeoffs between local communication (computer servers giving instructions to nearby mobile robots on obstacle avoidance) and global communication (computer servers sharing best quality solutions amongst each other). A scalable RADAR strategy in which the size of the computer servers (amount of memory and the number of mobile robots serviced by each of them) and the number of computer servers both increase as the system grows larger is shown to minimize communication times [26,27].

Peer-to-peer systems in which information is stored in a distributed fashion amongst thousands of nodes also have a tradeoff between local search within clusters of semantically similar nodes and global search between multiple clusters. A scalable RADAR strategy was shown to efficiently balance this tradeoff leading to fast search times [26].

We suggest that scalable RADAR principles are relevant for many other computing problems. In ongoing work, we are using scalable RADAR to design distributed intrusion detection systems with automated response.

Networks connect the components of complex systems. Natural selection has found network topologies that enable mechanisms for effective communication. Our work emphasizes the central role of network topology and interactions in how complex systems (living and artificial) distribute resources and information.

## VII. CONCLUSIONS

We posit that scalable RADAR requires more communication between individuals as systems grow. That requires that each individual have more opportunity to interact and more capacity for interaction. In the immune system we see that this is achieved in part by the semi-modular architecture of the lymphatic network. Larger animals have larger lymph nodes, and by virtue of being larger, they have greater ability to communicate through HEV with other lymph nodes in order to activate the larger required number of lymphocytes. As ant colony size increases, there is more communication between individual ants. This is facilitated by nest architectures in which larger colonies have higher densities of ants allowing for more interaction between ants. Additionally, larger colonies have greater diversity of ant castes, task groups and pheromones, thus there is more opportunity to encode information in interactions between different types of ants. Finally, pheromone signaling may be more effective in large colonies. It is essentially a mechanism to broadcast information, and that information reaches more ants as colony size increases. By using pheromones, each ant effectively increases its degree in the interaction network when the network is larger and denser.

These principles for scalable RADAR are relevant for engineered systems that use distributed search. In social networks, each node must communicate more for the system to operate effectively at large sizes. This requires innovations that allow longer distance communication and higher degree per node as the system grows. Scalable RADAR has been also shown to improve search times in peer-to-peer systems and multi-robot control systems. Scalable RADAR is useful not just for understanding how messages (or computer viruses or innovations) spread through social networks, but also for improving targeted searches in online domains, i.e. connecting the dots to find relevant related information in disparate sources. Finally, we suggest that innovations in communication are both cause and consequence of large networks, and as social network size grows, communication will continue to grow faster and more innovative to connect each individual to more other individuals.

Organisms, societies and computers are all complex systems whose behaviour emerges from the interactions of components. Our work emphasizes the pivotal role of network structure and interactions in understanding the emergent behaviour of living and artificial complex systems.

ACKNOWLEDGMENT

We acknowledge fruitful discussions with Dr. Stephanie Forrest, Dr. Frederick Koster and Dr. Alan Perelson. This work was supported by grants from NIH (P20 RR-018754) and DARPA (P-1070-113237).